\documentclass[runningheads]{llncs}
\usepackage{graphicx}
\usepackage{xcolor}
\usepackage{comment}
\usepackage{cite}
\usepackage[utf8]{inputenc}
\usepackage{multirow}
\usepackage{arydshln}
\usepackage{subfig}
\usepackage{tikz}
\usepackage{float}
\usepackage{flafter}
\usepackage{array}
\usepackage{tabularx}
\usepackage{graphicx}
\usepackage{url}
\usepackage{hyperref}
\graphicspath{ {./images/} }
\usepackage{amsmath, amssymb}
\usepackage{type1cm}        
\usepackage{makeidx}         
\usepackage{graphicx}        
\usepackage{multicol}        
\usepackage[bottom]{footmisc}

\usepackage{newtxtext}       %
\usepackage[varvw]{newtxmath} 
\usepackage{type1cm}        
\usepackage{makeidx}         
\usepackage{graphicx}        
\usepackage{multicol}        
\usepackage[bottom]{footmisc}

\usepackage{newtxtext}       %
\usepackage[varvw]{newtxmath}       

\makeindex 

\begin{document}

\title{Transfer Learning and Transformer Architecture for Financial Sentiment Analysis}

%
%

\author{Tohida Rehman\inst{1}\thanks{corresponding author} \and Raghubir Bose\inst{2}\thanks{corresponding author} \and Samiran Chattopadhyay \inst{1,3}\and\\ Debarshi Kumar Sanyal\inst{4}}

\authorrunning{Rehman and Bose et al.}
\titlerunning{Transfer Learning and Transformer Architecture for Financial Sentiment Analysis}
%

\institute{ Department of IT , Jadavpur University, Kolkata, India.\\
\email{tohida.rehman@gmail.com} \and  BFSI BTAG , Tata Consultancy Services.\\
\email{raghubir@gmail.com} \and  Institute for Advancing Itelligence(IAI) , TCG Crest; Jadavpur University, Kolkata, India. \\
\email{samirancju@gmail.com} \and Indian Association for the Cultivation of Science, Kolkata, India.\\
\email{dksanyal.india2@gmail.com}}
\maketitle 
\begin{abstract}
Financial sentiment analysis allows financial institutions like Banks and Insurance Companies to better manage the credit scoring of their customers in a better way. Financial domain uses specialized mechanisms which makes sentiment analysis difficult. In this paper, we propose a pre-trained language model which can help to solve this problem with fewer labelled data. We extend on the principles of Transfer learning and Transformation architecture principles and also take into consideration recent outbreak of pandemics like COVID. We apply the sentiment analysis to two different sets of data. We also take smaller training set and fine tune the same as part of the model\mbox{~}
\end{abstract}
.    
    
\section{Introduction}
Financial sentiment is a very important business indicator. Positive financial sentiments boost business while negative financial sentiment are catastrophic to business. This is because financial sentiment influences enterprises to stay ahead in business vision. It is understood that the definition of "\textbf{new and relevant information}" might change as updated information retrieval technologies become available and early-adoption of such technologies does provide an advantage. The sentiment analysis models trained on general aggregated dataset is not fully suitable in a financial domain. This is because financial texts have a specialized language coupled with unique vocabulary. They also have a tendency to use vague expressions instead of easily identified negative/positive words. Impact of Global pandemics like COVID on business sentiments and its cascading impact has become relevant. Approaches from Loughran and McDonald\unskip~\cite{Loughran} may seem a solution because they use "word counting" methods. The modern approach of efficient and effective use of transfer learning and transformation architecture methods appears to be more plausible solution to the challenges mentioned above, and are the focus of this paper.
    
\section{Objective}
The objective of this paper is to use an efficient mechanism to train language models with financial dataset. Once trained, the down-stream models are activated with the weights derived from the language modelling activity. This way it is possible to formulate a much better model for financial sentiment prediction. This approach ranging from  a single word embedding layer to the whole model, provides a solution to the scarcity of labelled data. Language models does not require any labels, as they are designed to predict the next word. Semantic representation is possible using Language models. Hence we need to fine-tune the labelled data so that the language model can use this information to predict the labels. Using transfer learning the language model can work on domain specific unlabelled corpus. Due to the contextual nature of the language used in financial domain, fine tuned context sensitive models are very effective.

The goal of this paper is to use fine-tune pre-trained language models for financial domain for sentiment analysis including the effect of COVID pandemic. To achieve the goal, the sentiment of a sentence from a financial news article will be tried to be predicted, using the Financial PhraseBank created by Malo et al. (2014) \cite{Malo} and FiQA sentiment scoring dataset \cite{Kiros} and the Worldbank COVID Financial sector response \cite{Erik}

The main objectives of this paper are the following. We evaluated the use multi-directional Transfer and Transformation model [BERT] on multiple financial sentiment analysis datasets and achieve a outperform modern sentiment scoring structure. This we intend to execute using a Pipeline based on BERT model. We also conduct several experiments to investigate catastrophic forgetting and hence fine-tuning without significant loss of performance.
    
\section{Related work and challenges}
\begin{figure*}[!htbp] 
\centering 
\includegraphics[width=11cm,height=7cm]{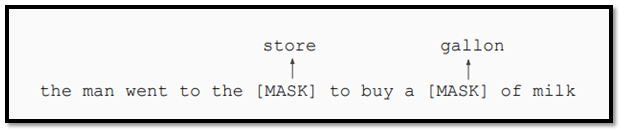}
\caption{Masked LM Scenario}
\centering
\label{fig:figure-f50cf4dc60d949a0bb3095472e72bea9}
\end{figure*}

The major deterrent for sentiment analysis in the financial domain is the lack of a substantial aggregated content of labelled dataset. This is the major reason for which it is not easy to apply proper neural network algorithms on the financial information available. The full potentiality of neural network is yet to be fully leveraged in financial domain as far as sentiment analysis is concerned. The problem of Language Models is that they mostly use a left context or a right context. 
Historically language model used the concept of end to end one way transfer to find the probability distribution and once trained words cannot reach themselves as in a bi-directional encoder. But now the natural language understanding is bi-directional that means it can capture the left context as well as the right context. The next effort is to have a masked language model or MaskedLM \unskip~\cite{Devlin}. We use a k\% masked data and try to predict the same by some means as seen below. 
The problem with the model below is too much training is required and not enough context may be available shown in figure \ref{fig:figure-f50cf4dc60d949a0bb3095472e72bea9}.

ELMo\unskip~\cite{Matthew}  or (Embeddings from Language Models) is one of the first successful research on modern approach where a pre-trained Model was used extensively . ELMo converts every input into its contextualized representation. Using the pre-trained weights of ELMo, contextualized word embeddings can be calculated for any word present as the input sentence. By initializing embeddings for down-stream tasks with any arbitrary input, were shown to improve performance on most tasks when compared to other approaches such as  GloVe\unskip~\cite{Robin}  or word2vec \unskip~\cite{Mikolov} . In ELMo learning from the first layer is not fully transferred to the next layer. ULMFiT uses novel approaches such as progressive unfreezing, slanted triangular learning rates and discriminative fine-tuning. ULMFit can be fine tuned to a whole pre-trained language model and can also be used for further pre-training a language model on a domain-specific aggregated data. This is based on the assumption that the target task information comes from a different source with respect to the general aggregated content used for training the model.
    
\section{BERT}
\begin{figure*}[!htbp] 
\centering 
\includegraphics[width=11cm,height=7cm]{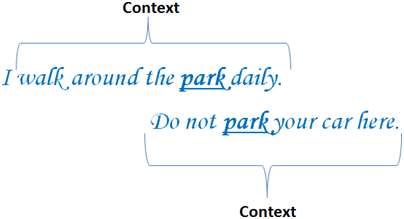}
\caption{Analyzing context of a sentence.}
\centering
\label{fig:figure-a22b31ed0afa4f54918a68eaca06c1f2}
\end{figure*}

Bidirectional Encoder Representations from Transformers (BERT) \cite{Devlin}  uses Bidirectional Transformers for language models. The key algorithmic segments of BERT is as follows:

  \begin{enumerate}
  \item \relax Pre-training from un-labelled texts.
  \item \relax Bi-Directional contextual models and transformation architecture
  \item \relax Text Entailment or Next Sentence Prediction.
  \end{enumerate}
  The key proposition of BERT is as follows:

BERT defines the activities of language modelling such as predicting randomly masked tokens in a sequence rater than the next token in addition to classifying sentences as following each other or not. BERT uses an efficient token based Masked Language Model. BERT uses a [CLS] token as a sequence approximate. The user can get the first token and use it for sequence prediction rather than a token prediction. BERT is a large neural network trained on an large aggregated data. To explain simpl, let us refer to the figure \ref{fig:figure-a22b31ed0afa4f54918a68eaca06c1f2}. 

\begin{figure*}[!htbp] 
\centering 
\includegraphics[width=11cm,height=7cm]{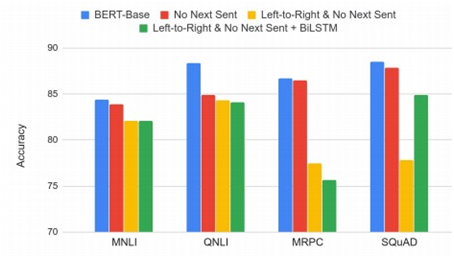}
\caption{Effect of BERT Pretraining}
\centering
\label{fig:figure-60e663c92b65452abfbb8abae6288c37}
\end{figure*}

If we try to predict the nature of the word ``park'' in the example above by only taking either the left or the right context, then we will have different views. One way to explain this better is to consider both the left and the right context before making a prediction. BERT can also be fine tuned to execute NLP tasks including natural language inference or question answering. How pre-training influences BERT is shown in figure \ref{fig:figure-60e663c92b65452abfbb8abae6288c37}. 

BERT uses the principle of Transfer learning \cite{Chuanqi}  as well.
  
\section{Implementation Methodology}
The overall methodology can be generically looked at the following steps.

  \begin{enumerate}
  \item \relax Setup the environment for the experiments. We will setup a virtual environment for the same. This includes setup of BERT models.
  \item \relax Bi-Directional contextual models and transformation architecture preparation of the dataset from various sources.
  \item \relax Training model is created. Classifier is run.
  \item \relax Prediction from an input data.
  \item \relax Conclusion from the prediction.
  \end{enumerate}

There are various controllable input parameters which are passed as configurations to the overall framework. The input parameters are then processed through the train of Bi-directional encoded transformers. The outcome of the work is a sentiment analysis score.
    
\section{Research Question and experimental setup}
This paper focusses on the following research questions

\begin{itemize}
  \item\relax Comparison of various pre-trained NLP models
  \item \relax Sentiment Analysis with discrete and continuous targets
  \item\relax How different training impacts model performance.
  \item\relax Selection of the right encoder
  \item\relax Sentence, Word and Character Embedding \cite{Reimers}
  \item\relax How many layers are introduced for model tuning.
  \item\relax The event of Catastrophic forgetting
\end{itemize}
  After the [CLS] token a dense layer is added. This is then applied on a labelled data-set for pre-training. The dataset for the paper is :

\begin{itemize}
  \item \relax TRC2-Financial. It is a collection of 1.8 million news articles published by Reuters.\cite{Lewis}
  \item \relax Financial PhraseBank : Made available by \cite{Malo}
  \item \relax COVID-19 Finance Sector Related Policy Responses : World Bank 2020\unskip~\cite{Erik}
\end{itemize}
  The Overall Model can be explained as below in figure \ref{fig:figure-39ffe6ae9a1a4268ad678fa8ec507713} : The stack of bi-directional encoders with masked LM is applied on the train of data till the sentiment position is reached. It has been seen \unskip~\cite{Howard}  that model performance for classification improves with fine tuning and additional pre-training. This is the reason why this paper has considered four set of pre-training data which is essentially derived from a large corpus of financial knowledge. The effect of training transfers across layers. One of the known risk that this paper intends to explore is the area of catastrophic forgetting and training the model to go past the same.

\begin{figure*}[!htbp] 
\centering 
\includegraphics[width=11cm,height=7cm]{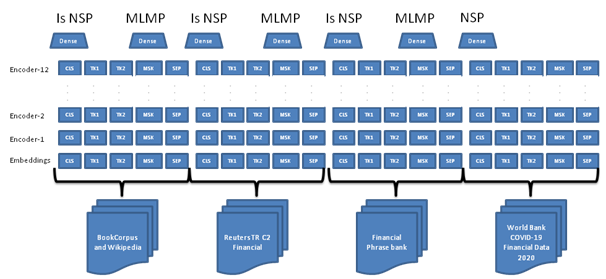}
\caption{BERT Process}
\centering
\label{fig:figure-39ffe6ae9a1a4268ad678fa8ec507713}
\end{figure*}

The Metrics that is considered in this paper are: accuracy, macro F1 average, loss due to Cross Entropy, mean Squared Error loss.  

In this paper we are considering the following :
\begin{itemize}
  \item \relax Warm-up proportion of 0.21
  \item \relax Dropout percent probability of 0.12 
  \item \relax Maximum sequence Length of 64 tokens
  \item \relax Learning rate of 2e - 5 
  \item \relax Mini Batch Size of 64 
  \item \relax Number of epoches is taken as 10 
  \item \relax Discrimination rate is 0.87. 
\end{itemize}
  The Environment Channels, dependencies are 

pytorch, anaconda, conda-forge - jupyter 1.0.0, pandas 0.23.4, python 3.7.3, numpy 1.16.3, nltk,  tqdm, ipykernel 5.1.3, pip: textblob, joblib 0.13.2, pytorch-pretrained-bert 0.6.2, scikit-learn 0.21.2, spacy 2.1.4, torch 1.1.0.

\begin{figure*}[!htbp] 
\centering 
\includegraphics[scale=0.7]{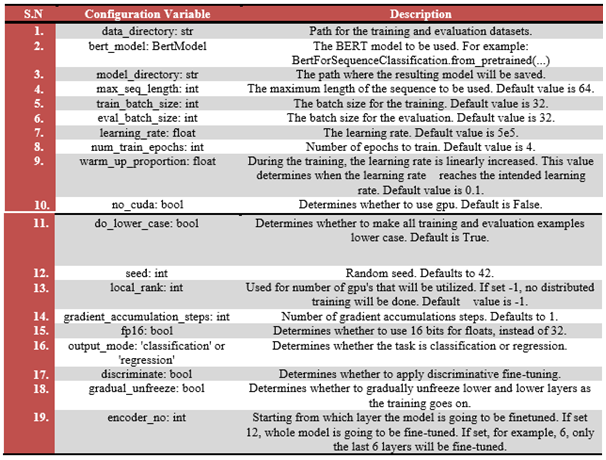}
\caption{{Configuration Data}}
\centering
\label{figure-9889de61e75047efaffc9697507d5953}
\end{figure*}

Google Cloud platform and Google Colab research project has been used for this scenario.The above models have been implemented in Pytorch have been tested on several data sets.The  performances matches with the associated TensorFlow implementations. This paper has used the following configurations for the model as shown in the figure \ref{figure-9889de61e75047efaffc9697507d5953}.

\begin{figure*}[!htbp] 
\centering 
\includegraphics[width=11cm,height=7cm]{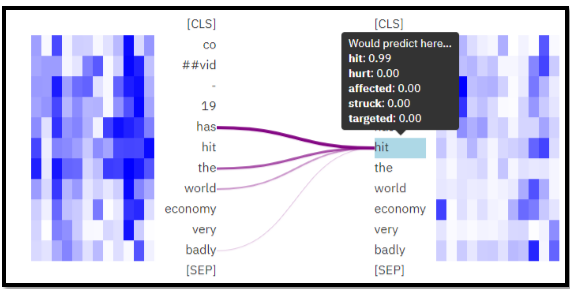}
\caption{Visual Analysis of BERT}
\centering
\label{fig:figure-40db4fb4c32b403c8dea496ed2668871}
\end{figure*}

\begin{figure*}[!htbp] 
\centering 
\includegraphics[width=12cm,height=8cm]{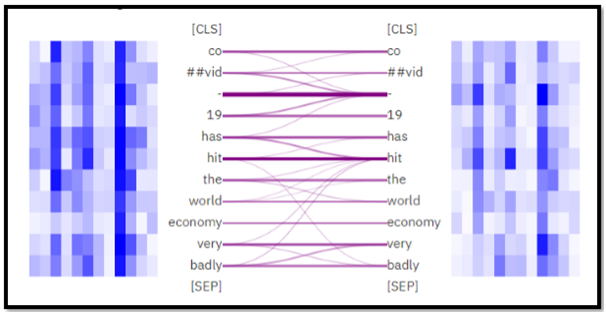}
\caption{Visual Analysis of BERT(12 Layers)}
\centering
\label{fig:b6d77225644644ef8b0b0035a82efe17}
\end{figure*}

The PreTrained Model which is used in BERT is used to store the configuration of the models. The pre-trained model also handles methods for model lifecycle management. In addition to the same also a few methods common to all models for the following purpose :

  \begin{enumerate}
  \item \relax   Resizing of the input embeddings. 
  \item \relax   Prune heads in the self-attention heads. 
  \end{enumerate}
There is also a Pre-trained configuration module available in BERT. It is the base configuration classes of BERT. The BERT configuration module is common to all models' configurations as well as life-cycle methods for managing the models. When the model is loaded with  a configuration file, the weights are not loaded. This process only affects the configuration of the model. The typical configuration of a BERT model is that with 12-layer along with 12-heads, 768-hidden and 110M parameters. Lower case English is used to train the model. We take the following text and try to visualize the BERT models. The input sentence being "COVID-19 has hit the world economy very badly",when subjected to a 12 layer configuration the corresponding visualizations of the BERT models can be seen from figure \ref{fig:figure-40db4fb4c32b403c8dea496ed2668871} and figure \ref{fig:b6d77225644644ef8b0b0035a82efe17}.

\section{Detailed Flow}

\begin{enumerate}
  \item \relax  The BERT object has a method called prepare\_model which sets up the structure in terms of processor of the dataset, cardinality of the labels, usage of GPUs, distributed training, steps of gradient accumulation and activity tokenizing.
  \item \relax The list of values of the labels in the dataset ie ['positive','negative','neutral']. 
  \item \relax The utility.py hosts all the utility classes used in the paper.  
  \item \relax  FinSentProcessor is responsible for processing all the data processing needs of the BERT object.
  \item \relax FinSentProcessor has the processing methods for the model.
  \item \relax The Usage of BertTokenizer.from\_pretrained function needs a discussion is to instantiate a PreTrainedBertModel from a pre-trained model file and download and cache the pre-trained model file if needed. 
  \item \relax The BertTokenizer includes tokenization of punctuation, splitting and wordpiece. 
  \item \relax The get\_data function gets the data for training or evaluation. It returns the data in the format that pytorch can use. 
  \item \relax  The create\_the\_model functioncreates the structure Sets the model to be trained and the optimizer.
  \item \relax Discriminative fine-tuning is used. 
  \item \relax BERTAdam is identified as an optimizer.
  \item \relax  The get\_Loader function creates a data loader object for a dataset.
  \item \relax The convert\_examples\_to\_features function is used to convert a data file into input feature list.
  \item \relax  The get\_metrics function is used to provide an endpoint for accuracy and precision-recall for different sentiments.
  \end{enumerate}

\section{Results}
\begin{figure*}[!htbp] 
\centering 
\includegraphics[width=11cm,height=7cm]{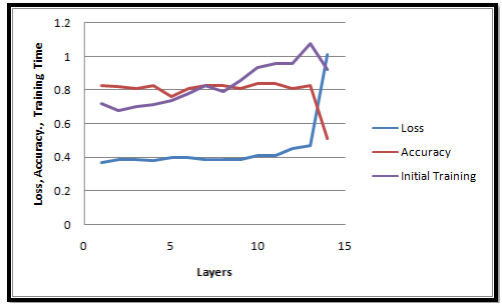}
\caption{Loss vs. Accuracy.}
\centering
\label{fig:figure-9ed696e4c1f94474bcaa30e837b35ad0}
\end{figure*}

The initial performance results obtained by training different layers BERT have 12 training Transformer encoder layers. At different depth loss and accuracy are different. This Loss-vs-accuracy for initial training can be seen in figure \ref{fig:figure-9ed696e4c1f94474bcaa30e837b35ad0}.

When this performance data is correlated between loss and accuracy we get the following correlation for BERT the results are shown below in figure \ref{fig:figure-4281637049e74006acd0ddd0d761314d}.

\begin{figure*}[!htbp] 
\centering 
\includegraphics[width=11cm,height=7cm]{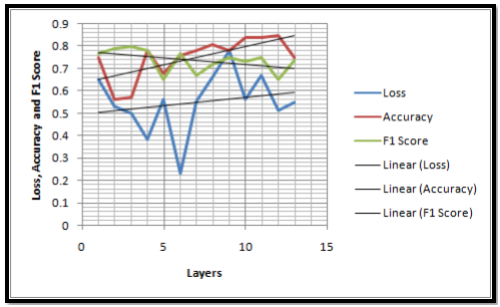}
\caption{Performance Data.}
\centering
\label{fig:figure-4281637049e74006acd0ddd0d761314d}
\end{figure*}

\section{Challenges }

  \begin{enumerate}
  \item \relax The problem with this approach using BERT is the size. The model used in BERT-base contains 110M parameters. There are 340M parameters in BERT-large. It is difficult to deploy a model of such size into many environments with limited resources, such as a mobile or embedded systems. 
  \item \relax At times BERT tends to overfit when subject to fine tuning.
  \item \relax Significant training and inference times.Training times does not impact main process. 
  \item \relax The inference times required by BERT vary depending on the model and needed hardware available but in many cases, this significantly limits us on quantity, cost and speed.  
  \item \relax Currently BERT language parameters are extremely large. Not all attention heads are always required.
  \item \relax There are known model limitations to BERT specifically when it comes to semantics and long term contexts. 
  \item \relax There is a high chance of Catastrophic forgetting.
  \item \relax The main challenge is interpretability of the transfer learning and transformation architecture - model driven interpretation is still a challenge. 
  \item \relax The combination of interpretability and transferability to downstream tasks is the major challenge that needs to be solved.
  \item \relax An adaptive and layered batch optimization technique  [LAMB] \unskip~\cite{Yang}  reduces BERT training time from 3 days to just 76 minutes.  
  \end{enumerate}

\section{Future Scope}
BERT based fine tuning has huge prospects in many areas of sentiment analysis in finance. In the area of finance there is a strong need to have a continuous sentiment analysis for any business data. The field of continuous fine-grained financial sentiment analysis will lead to a far more empathetic finance system which should be able to maximize human value. Further improvements can be done on a real-time stock market data and apply BERT based sentiment analysis to the same and predict the target price of the stock in a far accurate way. This way financial investment companies will be able to provide differentiated value to their end customers by minimizing the risk using right sentiment analysis technique using transfer learning and transformer architecture. The more the size and coverage of the Pre-trained data-set more in the accuracy and efficiency of the continuous sentiment analysis. There is a research work done in the area of combining the three different kinds of pretraining LM like bi-directional, left to right and seq-to-seq.

\bibliographystyle{IEEEtran}
\bibliography{ref}
\end{document}